\documentclass{AISB2008}
\usepackage{times}
\usepackage{graphicx}
\usepackage{latexsym}

\usepackage{multirow}

\title{\LARGE \bf
Robot-stated limitations but not intentions promote user assistance
}

\author{David Cameron\institute{Sheffield Robotics, University of Sheffield, UK, email: \{d.s.cameron, e.c.collins, jonathan.aitken, j.law\}@sheffield.ac.uk} \and Ee Jing Loh\institute{Department of Psychology, University of Sheffield, UK email: \{ejloh2, dxachua1\}@sheffield.ac.uk} \and Adriel Chua$^2$\and Emily Collins$^1$\and \\ Jonathan M. Aitken$^1$ \and James Law$^1$}

\begin{document}

\maketitle
\bibliographystyle{AISB2008}
\thispagestyle{empty}
\pagestyle{empty}

\begin{abstract}
Human-Robot-Interaction (HRI) research is typically built around the premise that the robot serves to assist a human in achieving a human-led goal or shared task. However, there are many circumstances during HRI in which a robot may need the assistance of a human in shared tasks or to achieve goals. We use the ROBO-GUIDE model as a case study, and insights from social psychology, to examine how a robot’s personality can impact on user cooperation. 

A study of 364 participants indicates that individuals may prefer to use likable social robots ahead of those designed to appear more capable; this outcome reflects known social decisions in human interpersonal relationships \cite{casciaro05}. This work further demonstrates the value of social psychology in developing social robots and exploring HRI.

\end{abstract}

\section{INTRODUCTION}

The use of autonomous, mobile service robots in the workplace is expected to grow substantially in the coming years \cite{hinds04}. Robots are anticipated to work side-by-side with people, assisting or collaborating with employees on a variety of tasks. As a consequence, Human-Robot Interaction (HRI) research typically explores interactions based around a robot in a supportive or assistive role for the human user \cite{Hancock11}. 

Current research in assistive robotics highlights the importance of ensuring users' trust that the robot can provide effective assistance. In scenarios of robot-assisted navigation, this may require: near-immediate user trust placed in a robot guide \cite{robinette2013building}; maintenance of appropriate levels of user trust, so neither over-reliance nor under-use occurs \cite{lee2004trust} \cite{desai2009creating}; and recovery of user trust after mistakes in automation \cite{mason2013trust, herrmann2013social}. Across the literature, user trust is typically associated with a robot establishing its capacity in meeting goals as an autonomous agent.

However, there are instances in HRI where an assistive robot guide may \textit{require} user support to achieve its aims or complete shared tasks. For example, a robot may hold incomplete information about its situation and require user input (e.g. asking for its location \cite{rosenthal10} or for directions \cite{bauer09}) or face a task requiring manual intervention by a user (e.g. autonomous mobile robots encountering physical barriers to progress \cite{Cameron15a,rosenthal10}). At this juncture, a robot cannot effectively demonstrate its capacity to operate as an autonomous agent and so typical channels for engendering user trust may be at risk. Robots requiring, rather than solely providing, user assistance in complex and social environments is an emerging topic \cite{Cameron15c} and so it becomes essential to study its impact on users' experience of HRI.

Research on the topic of robots requiring assistance identifies effective means for robots to determine their limits and when human intervention is required \cite{rosenthal10}, or how to locate users that can offer assistance  \cite{rosenthal12}. However, determining effective means of \textit{how} socially adaptive robots request help to encourage user cooperation and assistance is still a challenge in HRI.

To approach this challenge, we draw from social psychological models exploring cooperation between agents. Social psychological insights can be beneficial in exploring HRI, given it is a novel and still-developing research topic \cite{collins13}. Robots need to engage the user, particularly those that require user input (either directly through interacting with the robot or shaping the environment to meet the robot's needs). In particular, the emerging field of social robotics considers the optimization of a robot's morphology \cite{hinds04,breazeal99} and even its `personality' \cite{fernando14,cameron15b} as important to user perceptions of experienced HRI.

In this paper, we compare the impact of two different robot personalities on individuals' willingness to use a robot requiring assistance. We construct a friendly robot personality that empahsises its \textit{limitations} and a capable
personality that emphasises its \textit{intentions} in interactions. We further explore the impact that these personality features have on three factors identified in social psychology as having impact on human-human cooperation and assistance: \textit{liking, trust, and ambiguity}.  We test a model of robot personality; factors of liking, trust, and ambiguity; and individuals' willingness to help a robot requiring assistance in an applied HRI scenario.

\subsection{Pathways to cooperation}
The following paragraphs introduce the three factors antecedent to cooperation investigated in this experiment.  We target specifically the under-explored point of interaction: in which a robot needs human assistance, within the broader context of it operating as an assistive robot in a workplace environment.

\subsubsection{Liking}
Future robotic systems are considered likely to operate more as teammates rather than tools \cite{ososky2013building}. Understanding the impact this change could have on HRI is therefore vital to adapt work-forces to accommodate future, robotic teammates. Occupational psychology models of teamwork and cooperation in the workplace may provide a good foundation for understanding HRI with these robotic teammates. Research from occupational psychology indicates that individuals prefer to populate their cooperative working networks with people they like, ahead of those more capable in the cooperative role (but not liked) \cite{casciaro05,singh08}. Liking of individuals is considered to play a substantial role in motivations for cooperation and, particularly as the relationship develops, displace antecedent cognitive motivations for cooperation \cite{nicholson01}. 

\subsubsection{Trust}
A prominent model from occupational psychology of interpersonal cooperation in teammates identifies trust as its foundation \cite{mcallister95}. McAllister argues that to successfully achieve in a task requiring two agents working together, both need to trust each other. Specifically, cooperation is thought to benefit from both \emph{affective trust} (built by personable interactions from the partner) and \emph{cognitive trust} (built from evidence that one's partner carries out responsibilities competently) \cite{mcallister95}. Analogues for both forms are seen in HRI \cite{Hancock11}. A user's perception of a robot's performance (analogous to cognitive trust) and a user's perception of a robot's attributes, such as personality (analogous to affective trust), are found to positively contribute towards user trust in robots \cite{Hancock11}.

\subsubsection{Ambiguity}
Classic social psychology research indicates that ambiguity in assistive scenarios results in substantial detriment to individuals' proactive assistive behavior \cite{clark72}.  For many, HRI in cooperative environments may currently be entirely novel. As a result, individuals could face ambiguity in HRI situations and, without an indication of a robot's limitations, uncertainty regarding \textit{whether} the robot requires user cooperation. Alternatively, lack of a clear plan or intention communicated by a robot may create further ambiguity in \textit{how} the individual may best cooperate with the robot, limiting the action taken.

\section{Scenario}
To explore how a robot's personality may influence the above factors and user cooperation and assistance towards a robot, it is essential to consider an interactive social robotics scenario in which these circumstances arise. The ROBOtic GUidance and Interaction DEvelopment (ROBO-GUIDE) project \cite{law15} offers an ideal scenario to explore these factors. ROBO-GUIDE is implemented on the Pioneer LX mobile platform to autonomously navigate a multistory building, leading users (visitors to the building) to their chosen destination. 

Critically, there are elements in the tour which require user intervention to remove barriers to progress, such as using an elevator to navigate between floors. While the ROBO-GUIDE platform can identify which floor it is on \cite{McAree15}, it is currently unable to call for an elevator itself: it can neither manually operate elevator buttons nor remotely command elevator operation. As a result, the robot relies on user cooperation to press buttons to call the elevator and select the required floor to progress. To direct the user, statements of the robot's limitations or intentions are communicated using the on-board speech synthesizer.

Our focus for this study is the point at which the platform changes between floors as it navigates the multistory building. Use of ROBO-GUIDE requires users to place trust in an autonomous way-finding robot, whilst also assisting the robot in overcoming obstacles or barriers to progress. For both the user and the robot, this scenario is identified as a simple and low-risk circumstance in which a human user can act in order to meet a robot's needs \cite{Cameron15a}.

\subsection{Promoting user cooperation in HRI}
We identify how scenario-specific robot-stated limitations or intentions can influence factors for user cooperation and assistance.
The following sub-sections show proposed control-condition statements or requests in quotes and normal font, whereas supplemental, experimental statements are in the same quotes and italics.
\subsubsection{Liking}
ROBO-GUIDE's primary purpose is to lead new visitors to the robotics laboratory. It is anticipated that user liking of the robot would be greater with supplemental phrases regarding its offering `face-to-face' direct assistance to the human user. To promote \textit{liking}, we include friendly and relatable references by the robot that is assisting the user as a tour guide: ``Please follow me; \textit{I am here as your tour guide}''. 

\subsubsection{Trust}
It is anticipated that developing \textit{affective trust} overlaps substantially with developing user liking \cite{mcallister95} and can be built by demonstrating trust in others \cite{mayer95,Serva05}. To promote affective trust, we supplement requests from the robot for user cooperation at the elevator with identification of its limitations: ``Please press the down button; \textit{I can’t quite reach the buttons}''.

In contrast, \textit{cognitive trust} is developed through demonstration of an agent's competency in meeting its intended and/or required responsibilities \cite{mcallister95}. It is anticipated that users' cognitive trust of the robot's way-finding capabilities would be greater with the inclusion of additional phrases that directly address its intended aims: ``Please follow me \textit{to get to the robot labs}''.

\subsubsection{Ambiguity}
It is anticipated that many visitors would be unfamiliar with HRI or social robots and could find the experience unusual. Interacting with a novel robot, especially a robot that needs help, is anticipated to create ambiguous situations for users. To reduce ambiguity, we use friendly and relatable references by the robot about its limitations, which a user could help with in the HRI scenario: ``Please press ground floor; \textit{good thing you're here to do that for me}''.

It is further anticipated that ROBO-GUIDE's statements of intentions when requesting help will reduce ambiguity and promote cooperation. Without declaring why the tasks are to be completed, requests for help could be ambiguous in their purpose. Intentions should demonstrate the robot has a clear goal it is trying to achieve but that it is now facing an obstacle and so asks for help: ``Please press the down button \textit{to call the lift}'' and ``Please press ground floor \textit{for the Robot Labs}''. 

\subsubsection{A model of user cooperation}

\begin{figure}[thpb]
\centering
\includegraphics[width=3in]{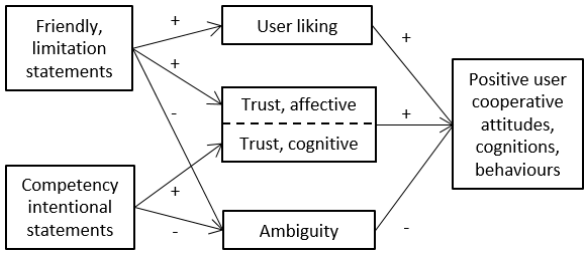}
\caption{Pathways by which robot-stated limitations and intentions impact on user-cooperation}
\label{Structure}
\end{figure}

We predict that a robot stating its limitations and intentions when requesting assistance will promote individuals' willingness to cooperate with assisting the robot. Pathways by which this is anticipated to occur, factors of liking, trust, and ambiguity, are over-viewed in Figure ~\ref{Structure}.

The robot stating its limitations is predicted to: promote user liking, promote users' affective trust towards the robot, and limit ambiguity in the situation. The robot stating its intentions is predicted to: promote users' cognitive trust towards the robot and limit ambiguity in the situation. These outcomes are in turn anticipated to positively impact on users' perceptions of assisting the robot

\section{METHOD}
\subsection{Design}

A 2x2 independent measures design was implemented. The four conditions comprised of the presence or absence of key statements regarding i) the robot's limitations, and ii) intentions. Participants were randomly allocated to a single condition and the Qualtrics survey engine prohibited repeat participation.

\subsection{Materials}

\subsubsection{Videos}

\begin{figure}[pb]
\centering
\includegraphics[width=3in]{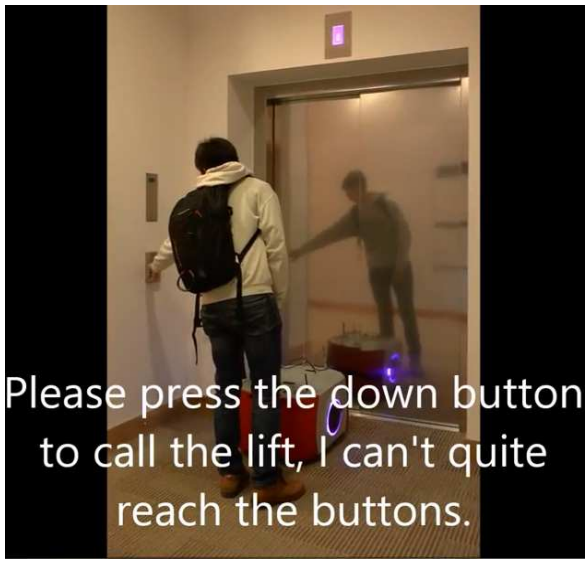}
\caption{Example of user assisting robot in HRI (Intentions and Limitations condition)}
\label{Video}
\end{figure}

Four HRI videos were prepared; one for each condition. The videos represented a typical use of the ROBO-GUIDE: an individual is greeted by the robot at the building entrance and instructed to follow it. The robot and user travel along a corridor to an elevator, at which point the robot instructs the user to press the call button (Example of HRI in Figure ~\ref{Video}). On entering the elevator, the robot again instructs the user to press the relevant button for the floor. The robot and user then leave the elevator at the target floor and travel along another corridor to the robot-lab - the final destination. 

All four films have minimal visual differences as non-critical scenes were used across conditions. Critical scenes differ in audio (i.e. the condition-specific words spoken by the robot) and subtitles of the robot's speech. Videos lasted 90 seconds.

\subsubsection{Questionnaires}

Perspectives on the observed HRI were assessed using the Godspeed Questionnaire \cite{bartneck09}. Measurements of liking were taken using the relevant Godspeed sub-scale. In addition, 3 ad-hoc measures were developed to assess participants' perspectives of the trustworthiness of ROBO-GUIDE, ambiguity of the interaction, and their likelihood of using ROBO-GUIDE. 

Trustworthiness was assessed using an ad-hoc 16-item scale derived from words previously identified as strongly associated with trust or mistrust in automated systems \cite{jian00}. Each of the following items was scored on a 7-point scale from \textit{not at all} to \textit{extremely}, headed `\textit{To what extent would you describe ROBO-GUIDE as...}' \textit{capable, competent, confident, deceptive, false, honest, honorable, incapable, incompetent, loyal, misleading, reliable, trustworthy, unreliable, unsure, untrustworthy}.

Ambiguity was assessed using a 4-item measure (alpha = 0.93) concerning the clarity of the robot's requests regarding, \textit{why}, \textit{whether}, \textit{when} and \textit{how} a user may need to assist the robot. Each item was scored on a 7-point scale from \textit{strongly agree} to \textit{strongly disagree}.

Likelihood of use was assessed using a 4-item measure (alpha = 0.78), with each item scored on a 7-point scale from \textit{strongly agree} to \textit{strongly disagree}. Items included: perceiving the robot as being fun to use, and intentions to use the robot in an unfamiliar building.

\subsection{Participants}

Participants were recruited through staff and student university volunteer mailing lists;  442 participants signed up for the study. A survey timer indicated that 78 participants did not fully watch the HRI video stimulus so were excluded from the study. Of the remaining 364 participants: 196 were female, 152 were male, and 16 declined to identify gender; 240 participants were British, 114 were from overseas, 10 declined to identify nationality; aged M = 27.67, SD = 8.99. Participants were given the opportunity to win one of two £50 gift vouchers as recompense for their time. 

\subsection{Procedure}

The survey was distributed through online mailing lists to reach a broad audience and delivered through the online survey engine Qualtrics (Version 06.2015; Copyright 2015; Qualtrics, Utah). Participants were first presented with a study information and consent page. On agreeing to participate, individuals were randomly allocated to one of the four conditions described above and presented with an HRI video. Following the video, participants completed the Godspeed questionnaire, measures rating trustworthiness of ROBO-GUIDE, interaction ambiguity, and willingness to use the ROBO-GUIDE. At the close of the questionnaire, participants could share demographic details and register for a chance to win offered gift vouchers. The study took participants approximately 10 minutes to complete.

\section{RESULTS}

\subsection{Randomisation Check}

Participant numbers were evenly distributed across conditions \textit{x$^{2}$}(1, N = 364) =  0.01, \textit{p} = 0.92. For each condition, there were even distributions of participants in terms of gender and nationality (max \textit{x$^{2}$}(1, N = 354) = 3.31, \textit{p} = 0.07).

\subsection{Primary results}
\subsubsection{Liking}

There was a significant main effect for the robot stating its limitations \textit{F}(1,356) = 53.407, \textit{p} $<$ 0.01. Participants who saw the interactions in which the robot stated its limitations reported liking the robot to a substantially greater extent (M = 3.77, S.E. = 0.05) than those who saw control statements (M = 3.24, S.E. = 0.05). This is a large effect observed (d = 0.77). There was also a significant main effect for the robot stating its intentions \textit{F}(1,356) = 7.25, \textit{p} $<$ 0.01. Participants who saw the interactions in which the robot stated its intentions reported liking the robot to a substantially greater extent (M = 3.60, S.E. = 0.05) than those who saw control statements (M = 3.40, S.E. = 0.05). This is a small effect observed (d = 0.28). There was no significant interaction effect between the robot stating both its limitations and intentions on participants' liking \textit{F}(1,356) = 0.87, \textit{p} = 0.79. 

\subsubsection{Trust}

The 16 items for trust were subjected to a principal axis factor analysis for the full sample of the 364 participants; missing values were treated pairwise. Three factors with Eigenvalues greater than 1.00 \cite{cattell66} were extracted from the matrix, explaining 59\% of the variance. Inspection of the pattern matrix showed that all items loaded above 0.40 on one of the three factors. Items on the first factor are mainly related to falsity: \textit{deceptive, false, misleading, unreliable, unsure}, and \textit{untrustworthy}. Items on the second factor are mainly concerned with affective trust: \textit{honest, honorable, loyal}, and \textit{trustworthy}. Finally, items on the third factor are mainly concerned with cognitive trust \textit{capable, competent, confident, reliable}, and two negatively scored items of \textit{incapable} and \textit{incompetent}.

However, there were no significant main effects for the robot stating its limitations nor intentions for either factors of affective trust (Limitations \textit{F}(1,351) = .11, \textit{p} = .75; Intentions \textit{F}(1,351) = .21, \textit{p} = .65) or cognitive trust (Limitations \textit{F}(1,351) = 2.14, \textit{p} = 0.14; Intentions \textit{F}(1,351 = .01, \textit{p} = .98). Furthermore, there were no significant interaction effects.

\subsubsection{Ambiguity}

There was a significant main effect for the robot stating its limitations \textit{F}(1,352) = 134.95, \textit{p} $<$ 0.01. Participants who saw the interactions in which the robot stated its limitations reported substantially less ambiguity in the situation (M = 1.87, S.E. = 0.10) than those who saw control statements (M = 3.52, S.E. = 0.10). This is a large effect observed (d = 1.22). There was no main effect for the robot stating its intentions on participants' perceptions of ambiguity in the interaction \textit{F}(1,352) = 0.43, \textit{p} = 0.52.

\begin{figure}[htbp]
\centering
\includegraphics[width = 3in]{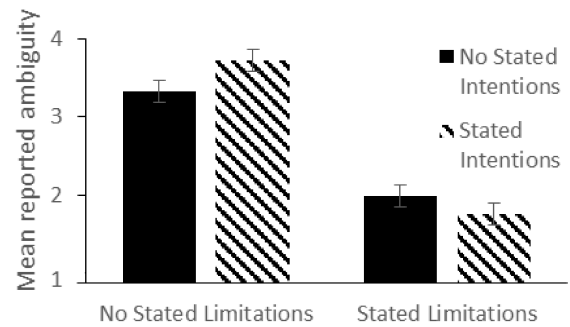}
\caption{Interaction effects for intentions and limitations for user perceived ambiguity}
\label{interaction}
\end{figure}

There was a significant interaction effect between the robot stating both its limitations and its intentions \textit{F}(1,352) = 4.72, \textit{p} = 0.03. The robot solely stating its intentions promoted greater perceptions of ambiguity in comparison to control statements (M = 3.72, SE = 0.14 versus M = 3.32, SE = 0.14). However, in combination with the robot stating its limitations, participants' ratings of ambiguity were lower in comparison to the robot solely stating limitations only (M = 1.76 SE = 0.14 versus M = 1.98, SE = 0.14). Simple main effects analysis showed that the robot-stated intentions resulted in more ambiguity than control statements, when presented without robot-stated limitations (p = .04), but there were no differences when accompanied with robot-stated limitations (p = .28). These results are presented in Figure ~\ref{interaction}.

\subsubsection{Willingness to use}

There was a significant main effect for the robot stating its limitations \textit{f}(1,352) = 8.74, \textit{p} $<$ 0.05. Participants who saw the interactions in which the robot stated its limitations reported greater willingness to use the robot (M = 4.57, SE = 0.11) than those who saw control statements (M = 4.26, SE = 0.11). There was no significant main effect for the robot stating its intentions on participants' willingness to use the robot \textit{F}(1,352) = 1.07, \textit{p} = 0.30. There was no significant interaction effect between the robot stating both its limitations and intentions on participants' willingness to use the robot \textit{F}(1,352) = 0.09, \textit{p} = 0.86.

\begin{figure}[htpb]
\centering
\includegraphics[width = 3in]{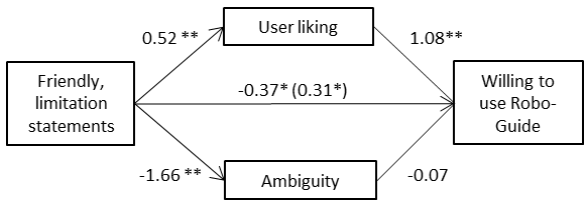}
\caption{Regression coefficients for the relationship between robot-stated limitations and individuals' willingness to use ROBO-GUIDE as mediated by user liking and ambiguity. The regression coefficient between robot stated limitations and willingness to use ROBO-GUIDE, controlling for user liking, is in parenthesis (* p $<$ 0.5; ** p $<$ 0.01)}
\label{Structure2}
\end{figure}

Mediation analysis was run to determine if the main effect of the robot-stated limitations promoting individuals' willingness to use ROBO-GUIDE could be explained by the effects seen above for robot-stated limitations on user liking and ambiguity\footnote{User trust is not included as there were no main effects observed.}. Indirect effects were computed following 1,000 bootstrapped samples at 95\% confidence intervals. The positive relationship between robot-stated limitations and individuals' willingness to use ROBO-GUIDE was mediated by user liking but not ambiguity (see Table ~\ref{table}). Figure ~\ref{Structure2} identifies the pathways examined in the mediation analysis and coefficients between constructs.
 
\begin{table}[htbp]
\caption{Mediators for robot-stated performance limitations on individuals' willingness to use ROBO-GUIDE}
\label{table}
\begin{center}
\begin{tabular}{c c c c c}
\hline
\multirow{2}{1.7cm}{\centering Dependent variable} & Mediators &  \multirow{2}{1.7cm}{\centering Point estimate effects} &	\multicolumn{2}{c}{\multirow{2}{2cm}{\centering 95\% confidence intervals}}\\
& & & \multicolumn{2}{c}{} \\
& & & Lower & Upper\\
\hline
Willingness & Liking & 0.56 & 0.39 & 0.74 \\
to Use & Ambiguity & 0.12 & -0.04 & 0.031 \\
& Total & 0.31 & 0.01 & 0.62 \\
\\
\multicolumn{5}{c}{R$^{2}$ = 0.27, \textit{F}(3,352) = 42.54, p $<$ 0.01} \\ 
\hline
\end{tabular}
\end{center}
\end{table}

\subsection{Demographics}
There were significant main effects of nationality for both participants' reports of liking the robot \textit{F}(1,343) = 6.03, \textit{p} = 0.02 and their perceptions of the robot as being trustworthy \textit{F}(1,342) = 21.68, \textit{p} $<$ 0.01. Individuals from overseas reported liking the robot more (M = 3.66, SE = 0.07) than those from the UK did (M = 3.45, SE = 0.05) and showed more affective trust towards the robot (M = 4.50, SE = 0.12), than those from the UK did (M = 3.81, SE = 0.09). These are small and medium effects (d = 0.28; d = 0.53) respectively. There were no significant main effects for gender (Max \textit{F}(1, 354) = 2.58, \textit{p} = 0.11).

There were significant interaction effects between nationality and gender for both perceptions of robot competency \textit{F}(1, 343) = 5.83, \textit{p} = 0.02 and willingness to use such a robot \textit{F}(1,343) = 7.80, \textit{p}  0.01. In both cases, female participants from the UK scored higher (M = 5.31, SE = 0.08; M = 3.96, SE = 0.13) than female participants from overseas (M = 4.94, SE= 0.13; M = 3.60, SE = 0.21), respectively; whereas, male participants from overseas scored higher (M = 5.26, SE = 0.12; M = 4.12, SE = 0.20) than male participants from the UK (M = 5.11, SE = 0.10; M = 3.48, SE = 0.16), respectively.

\section{Discussion}
The results demonstrate a substantial difference in individuals' responses to observing a friendly versus capable social robot. The contrast between a friendly, limitations-focused, and capable, intentions-focused, robot is seen in individuals' willingness to use the robot in the future; only the former significantly differs from the control condition. Moreover, the observed effect can be explained by the mediating influence of individuals liking a limitations-focused robot. As observed in occupational psychology \cite{casciaro05}, individuals prefer to further interact with those they like, rather than those presented as being more capable. This study provides new evidence that strategies used by individuals in interpersonal relationship development can be extended to apply to social robotics HRI. 

One key strength of the current work is the available sample size, which would require an extensive effort to produce in a field robotics study. The large sample size enabled a sufficiently powered 2x2 design study and a full factor analysis of the ad-hoc trust scale. Again, the factor analysis indicates that it is extremely worthwhile exploring HRI in terms of social psychological models of interpersonal relationships. Factors extracted from the user responses closely corresponded to the constructs of affective and cognitive trust \cite{mcallister95}. Participants may be applying their own understanding of human-human social relationships to novel contexts (HRI) containing social agents.

There are several outcomes apparent due to individual differences, such as an interaction effect of gender and nationality on willingness to use ROBO-GUIDE. This emphasises the importance of human-focused study design and robotic development \cite{Cameron15c}. The demographic differences in individuals' responses indicate that developing a socially adaptive robot responsive to users, rather than attempting a `one-size-fits-all' approach could greatly benefit user perceptions of HRI. In terms of application to the ROBO-GUIDE scenario, this could comprise of adjusting synthetic personalities following a user's feedback in preparation for their next use of ROBO-GUIDE.
\subsection{Further Directions}
\addtolength{\textheight}{-10cm}
The promising findings so far offer many directions to further examine user assistance in HRI.

The scenario chosen demonstrates a user assisting a robot at minimal cost to themselves. While the explored factor of ambiguity is important in determining proactive assistive behaviour \cite{clark72}, relative cost or risk to the individual is also a key factor \cite{piliavin75}. The present work indicates the two robot personalities both reduce ambiguity in interaction. Alternative scenarios (such as in manufacturing), in which a user must expend greater effort to assist a robot, may further see changes in the impact of ambiguity on a user's willingness to assist.

A complex social interaction between user and robot, such as the scenario presented, can be further explored across a variety of social dimensions and factors not-yet considered. Empathy towards robots in need or at risk \cite{rosenthal2013neural} may contribute to user willingness to assist. This may impact on the current pathways identified, particularly towards a friendly robot or for individuals more likely to anthropomorphise the robot \cite{riek2009anthropomorphism}. Alternatively, the interaction could be further considered in a broader, social context, rather than the applied, occupational psychology background for this workplace-inspired interaction scenario and draw from related social cognition literature \cite{fiske2007universal}. 

Further revisions to the scenario could include altering the robot's behavioural competency. It is possible that the robot performing effectively throughout the task (elevator operation not-withstanding) places a ceiling on individuals' trust towards the robot and limits the impact of the experimental conditions. As identified in the literature \cite{Hancock11}, errors from the robot impinge on user trust. Introductions of errors (such as departing the elevator on the wrong floor) may lower baseline user trust, enabling any potential influence from limitation or intentional statements. This could further develop work examining the impact of robots addressing their mistakes on improving user HRI experiences \cite{snijders15}, again suggesting the importance of user liking ahead of competency in social robots. 

This study forms a solid foundation to conduct a field experiment exploring user behaviour in the HRI scenario. Direct measures of user behaviour, such as human-robot `interpersonal' distance \cite{walters09} and user speed of response to robot requests, would build on the established results of willingness to engage, liking of, and ambiguity felt towards the social robot.

Our findings have implications for the development of social robots and study of social HRI. The study demonstrates that individuals may apply known human-human interpersonal social decisions to social robots. The careful development of robot personalities to account for this may be instrumental in fostering positive user experiences of social robots and effective HRI.


\ack
The research team thank Man Tik Cheung (Hugo) for his contribution to the video materials used in the study. This work was supported by European Union Seventh Framework Programme (FP7-ICT-2013-10) under grant agreement no. 611971 and the University of Sheffield's SURE scheme
\bibliography{bibfile}
\end{document}